\documentclass[runningheads]{llncs}
\usepackage{graphicx}

\begin{document}
\title{MRI to FDG-PET: Cross-Modal Synthesis Using 3D U-Net For Multi-Modal Alzheimer's Classification}
\titlerunning{MRI to FDG-PET}
%

\author{Apoorva Sikka, Skand Vishwanath Peri, Deepti R. Bathula}
\institute{Indian Institute of Technology Ropar\\
%
\email{apoorva.sikka@iitrpr.ac.in, pvskand@gmail.com, bathula@iitrpr.ac.in }\\
}

\maketitle              
\begin{abstract}
 Recent studies suggest that combined analysis of Magnetic resonance imaging~(MRI) that measures brain atrophy and positron emission tomography~(PET) that quantifies hypo-metabolism provides improved accuracy in diagnosing Alzheimer's disease. However, such techniques are limited by the availability of corresponding scans of each modality. Current work focuses on a cross-modal approach to estimate FDG-PET scans for the given MR scans using a 3D U-Net architecture. The use of the complete MR image instead of a local patch based approach helps in capturing non-local and non-linear correlations between MRI and PET modalities. The quality of the estimated PET scans is measured using quantitative metrics such as MAE, PSNR and SSIM. The efficacy of the proposed method is evaluated in the context of Alzheimer's disease classification. 
The accuracy using only MRI is 70.18\% while joint classification using synthesized PET and MRI is 74.43\% with a p-value of $0.06$. The significant improvement in diagnosis demonstrates the utility of the synthesized PET scans for multi-modal analysis.
\end{abstract}

\section{Introduction}
\label{sec:intro}
Alzheimer's disease~(AD) is a chronic neuro-degenerative disorder that causes problems with memory, thinking and behavior. It is a progressive disease that gets worse with time, making early diagnosis very crucial  \cite{adni}. Recently, various techniques using multi-modal image analysis have been proposed to identify bio-markers that aid in accurate diagnosis of AD \cite{multimodal}. It is evident that multiple modalities provide complementary information related to the disease, which when combined increases the efficacy of diagnosis. Joint analysis of positron emission tomography~(PET) and magnetic resonance imaging~(MRI) has been accepted as a method to diagnose AD \cite{ClinicalEvidence}. While gray matter atrophy and ventricular enlargement in MRI are established markers for pathology, pattern of neuronal uptake and cerebral distribution of FDG in PET is also a discriminating factor for AD. However, PET in comparison to MRI is a relatively new modality and acquiring different modality scans for a single patient is not always feasible due to high cost, lack of imaging facilities and increased risk of radiation exposure. Current work attempts to use the information from MR image to estimate virtual PET scan and further explores the value of these synthetic PET scans in enhancing disease prediction accuracy when combined with MRI.

In recent years, various approaches based on machine learning have been proposed to predict one modality from another or a combination of other modalities. To reduce radiation dose, \cite{ACM-GAN} employs context-aware GANs to predict CT scans from MR scans. Similarly, a regression forest based framework was developed in \cite{lowdose} for predicting a standard-dose brain FDG-PET from a low-dose PET image and its corresponding MRI. A more challenging cross-modal synthesis task involves predicting functional scans from their corresponding structural scans. \cite{patch} used a patch based CNN which is capable of capturing non-linear mappings between PET and MRI. Few techniques based on partial least squares regression~(PLSR) \cite{multimodal} or independent component analysis~(ICA) \cite{ICA}, mapping non-local correlations have also been proposed.

\begin{figure*}[!h]
  \centering
  \includegraphics[width=0.47\textwidth,angle=180]{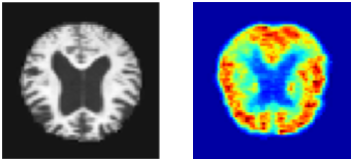}
  \caption{A sample pair of MRI and its corresponding PET slice depict lack of one-to-one correspondence at individual voxels or local neighborhoods in terms of signal intensities and details.}
   \label{fig:intro}
\end{figure*}

There are two important aspects to consider when estimating FDG-PET images from their corresponding MRI Scans: (a) correlation between these modalities is not purely local or one-to-one; PET is a functional modality that quantifies hypo-metabolism and MR being a structural one measures brain atrophy \cite{jack1}, \cite{CM} and (b) relationship between these modalities is quite complicated and nonlinear. The Figure \ref{fig:intro} shows corresponding slices of MR and PET scans of the same subject where the correlation values between local patches vary significantly in the range $[-0.86, 0.94]$. The technique in \cite{patch} based on deep learning uses patches that are defined in a local manner where a local patch of MR corresponds to a local patch of PET to learn the mapping. In contrast, the PLSR method \cite{multimodal} tries to capture global or non-overlapping correlations in the images, but maps inputs and outputs in a linear fashion. To effectively estimate PET from MR, the model has to capture relationships between non-adjacent voxels and learn non-linear mapping from MR to PET.

Deep learning architectures have recently emerged as the most prominent and powerful techniques to learn complex non-linear transformations. Specifically, convolution neural networks~(CNN) have been used to solve various problems like image recognition, segmentation and classification in computer vision. These models are hierarchical and have proved to be able to extract different types of features automatically. 
Unlike patch based networks that involve extracting and training using millions of patches (making them computationally inefficient), these networks directly process the entire input image to generate the final output image.


One such recently used network is deep convolutional auto-encoders based on U-Net architecture \cite{U-Net} with skip-like connections. In this work, conventionally unsupervised auto-encoders \cite{fully_autoencoders} are used in a supervised setting to learn a joint mapping between the two-modalities. But a major drawback for fully-connected networks is the increased chance of overfitting due to very large number of parameters and less training data available in medical images. Consequently, 3D fully convolutional network is employed that is not only capable of capturing global / non-local correlations but also models the input-output mapping using non-linear functions. The loss function used is binary cross-entropy as the loss function. As the accuracy of the predicted PET scans determines the success of any type of downstream analysis (classification, ROI analysis, etc.), the quality of the synthesized PET scans is thoroughly evaluated using quantitative metrics against ground truth. We have used three metrics namely, mean-absolute loss~(MAE), peak signal-to-noise ratio~(PSNR) and structural similarity index~(SSIM). The efficacy of the proposed approach is further evaluated through multi-modal AD classification using logistic regression.

Our contributions in this paper are the following:
\begin{itemize}
\item We propose the first global and non-linear cross-modal approach for PET estimation from MR images via adapting 3D convolutional U-Net\cite{3DUNET} architecture which takes care of non-local intensity correlations as well as non-linear mapping of input to output.
\item We extensively evaluate our proposed cross-modal method against the existing patch based estimation method on three different metrics.
\item We further assess the significance of estimated PET scans from the proposed method on the task of multi-modal Alzheimer's Disease Classification.
\end{itemize}

\section{Methods}
Given T1 weighted MRI scans represented as $X$ and FDG-PET scans $Y$ for $k$ sample subjects with size $x_1 \times x_2 \times x_3$ and $y_1 \times y_2 \times y_3$ respectively. The task is to learn a mapping between the above mentioned modalities where every voxel in the given input scan is used to predict every voxel in the output scan. We use a 3D U-Net based architecture to estimate the corresponding PET scan. We have used gray matter~(GM) from MRI scan as an input to the model. Gray Matter is used to estimate PET scan due to its high correlation in determining AD and normals. The architecture though is not a fully connected network but repetitive convolutions at multiple layers added with skip connections ensures that global correlations are captured.    

\begin{figure*}[t]
  \centering
  \includegraphics[width=0.99\textwidth]{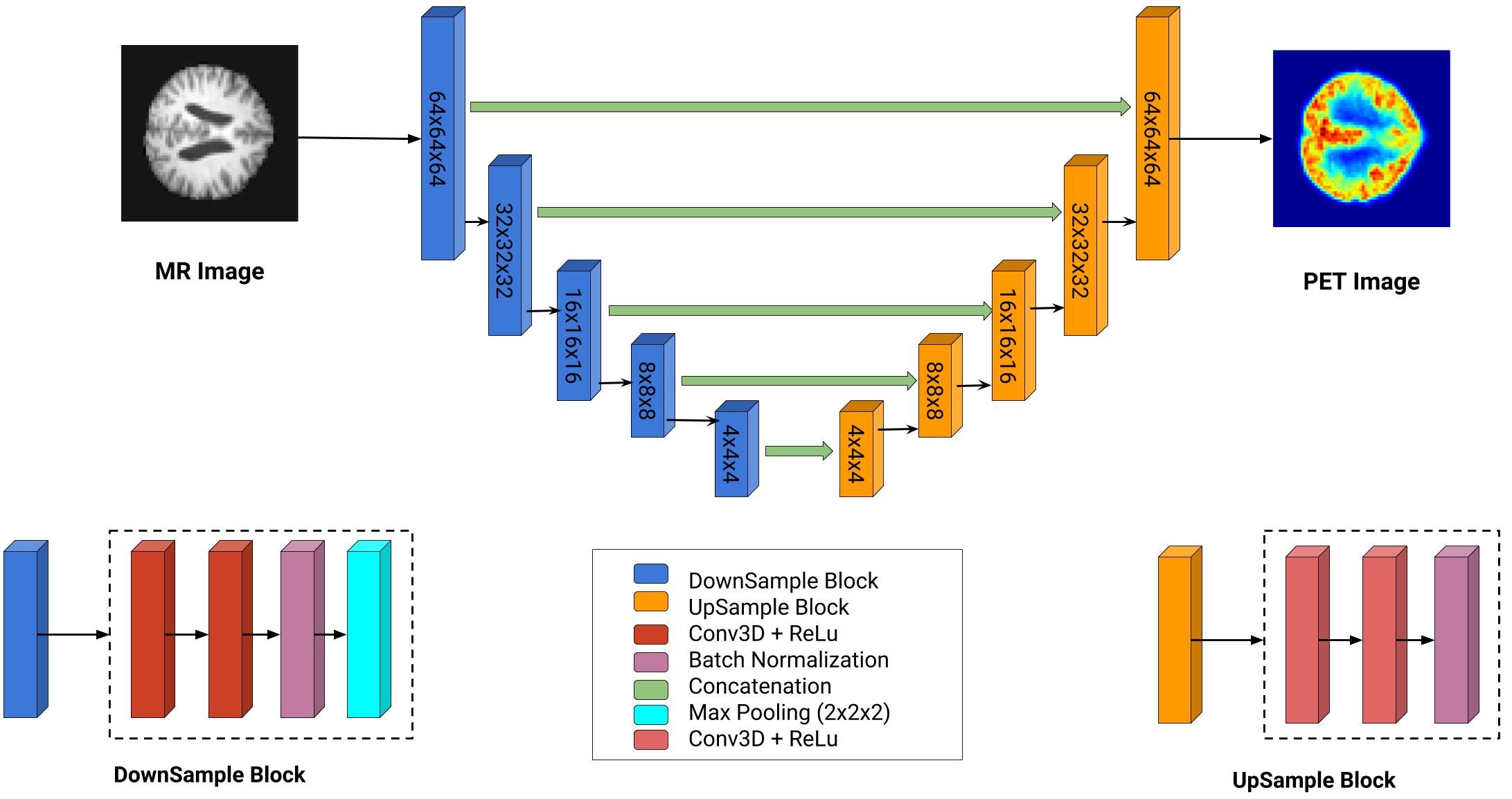}
  \caption{3D Regression U-Net architecture to generate a virtual PET scan using MRI scan.}
   \label{fig:wflow}
\end{figure*}

\subsection{Dataset}
The experiments have been conducted on scans obtained from Alzheimer's Disease Neuroimaging Initiative~(ADNI) database ~(adni.loni.usc.edu) \cite{adni}. A total of 384 subjects having both MRI and PET modalities~(192 Normals and 192 AD) were used. We chose all the subjects that had both MR and PET images from the ADNI dataset available online. This includes scans from different populations from different areas as mentioned in \cite{demographicsADNI}. Since we don’t use any clinical score or demographic details as input or as ground truth in our method we did not mention details in the paper.

\subsection{Preprocessing}
As the downloaded ADNI dataset was not pre-processed, we implemented a manual pre-processing pipeline. MR Scans were first registered to MNI template space using FSL package \cite{fsl}.This was mandatory since all the images are of different sizes and registering them to a common space is required (to maintain homogeneity). All the images can be registered to a common image but we chose MNI since it is a standard template in the literature as mentioned in \cite{lakshmanan}. The registered scans were then skull-stripped using ROBEX \cite{robex} which were further normalized to $[0-1]$ intensity values (to avoid the problem of gradients exploding and convergence). These preprocessed scans were then segmented into gray matter~(GM), white matter~(WM) and cerebral-spinal fluid~(CSF) using FSL. There were multiple PET images for the same subject. So, we first registered all the other PET images to one of the reference PET and then we averaged all the registered PET images. These averaged out PET images were aligned to MR images followed by registration to the same template space and then normalized to $[0-1]$ intensity values. Both MR and PET scans were then down sampled to $64 \times 64 \times 64$ voxels to reduce the network parameters (Figure \ref{fig:example} clearly shows that although the size was small the details in the images were still intact and we could figure out the difference between AD and normals. This was a trade off between computational power and size of the image.)

\subsection{Network Architecture}
Figure \ref{fig:wflow} shows the complete architecture of the modified network. It is similar to 3D U-Net \cite{U-Net} architecture but has a sigmoid layer as the last layer which generates the estimated PET image with same dimensions as the input MR image. In this network, the weights are optimized to perform a regression task, where the objective is to learn a mapping between gray matter extracted from MRI scans and PET scans.

The network essentially comprises of an encoder and a decoder. Encoder is a modified convolutional network. It consists of $3 \times 3 \times 3$ convolutions followed by a rectified linear unit and a down-sampling operation performed using $2 \times 2 \times 2$ max pooling with a stride of $2$. With each layer, we have increased the number of features maps. The decoder is an expanding network where every step consist of an upsampling followed by $3 \times 3 \times 3$ convolutions and batch normalization layers. In addition to the simple encoder-decoder architecture, there are skip connections which concatenate features of corresponding size from encoder to decoder layers. Although the inclusion of down-sampling layers leads to loss of information, the addition of skip connections between enocder and decoder helps retain some amount of information across higher layers from lower layers. This leads to generation of slightly smoother scans in comparison to ground truth. The final layer is $1 \times 1 \times 1$ convolution consisting of sigmoid activation which brings all values to $[0-1]$ pixel range. Weights of the network are optimized in an end to end manner using binary cross entropy loss function as shown in equation: 
\begin{equation}
H(y_{t},\hat{y_t})= - \sum_{x}y_{t}(x) \; log \; \hat{y_t} \\ (x)
\end{equation}
where  where $\hat{y_t}$ is the predicted PET scan, $y_{t}$ the ground truth PET scan and $x$ is the total number of samples present in a batch.




\section{Experiments}
The proposed architecture was trained on NVIDIA Geforce GTX1080. For efficient use of the available data, 9-fold cross-validation was performed where $1$ fold was used as a test set, $1$ fold for validation and remaining $7$ folds were used as training data. The learning rate was set to $0.008$ and Adam \cite{adam} optimizer was used. The architecture is shown in Figure \ref{fig:wflow} where number of feature maps of the model were chosen according to the GPU RAM. Total number of trainable parameters were 20M which is less as compared to a fully connected network and fully convolutional network \cite{fcn} of similar size. U-Net introduces skip connections which enhances learning by merging various lower layer features to higher layers. The model was trained for 10 epochs.

The proposed method is evaluated against another comparative method by Li et. al \cite{patch} that models the relationship between MRI and PET in a local, non-linear fashion. Firstly, preprocessed images were used to extract Gray Matter scans. Then, 3D patches from these images of size $15 \times 15 \times 15$ were extracted and the corresponding patch for the PET image of size $3 \times 3 \times 3$ is reconstructed. The method published in \cite{patch} was replicated to the maximum extent possible based on details provided in the paper, as the code is not available. All the parameters were kept as mentioned in the paper.
The following section depicts the efficacy of the method by evaluating estimated PET scans both quantitatively and qualitatively. Quantitative evaluation is done using three different metrics based on correlation, perception and pixel intensities.
\vspace{-3mm}
\subsection{Mean Absolute Error~(MAE)}
MAE is a commonly used metric for any reconstruction problem. It gives us the average absolute difference between the estimated image and the ground truth intensity values. It is computed as follows:
\begin{equation}
MAE = \frac{\sum_{i=1}^{n}{|y_i - x_i|}}{n}
\end{equation}
$x_{i}$ and $y_{i}$ are the intensity values of the pixels of estimated and actual PET.	

\subsection{Peak Signal-to-Noise Ratio~(PSNR)}
PSNR is mostly used as a quality measurement between two images. PSNR represents a measure of peak error. It is computed as follows:
\begin{equation}
PSNR= 10 \; log_{10} \Bigg( \frac{MAX^2}{MSE} \Bigg)
\end{equation}
where $MAX$ is the maximum possible intensity of the image and $MSE$ is the mean squared error between the estimated and ground truth PET image.

\subsection{Structure Similarity Index~(SSIM)}
Unlike MAE which measures the quality of an image based on pixel intensities, SSIM compares the similarity in structures of the two images. It is computed using the following equation:
\begin{equation}
SSIM(x,y) = \frac{(2\mu_x\mu_y + C_1) + (2 \sigma _{xy} + C_2)} 
    {(\mu_x^2 + \mu_y^2+C_1) (\sigma_x^2 + \sigma_y^2+C_2)}
\end{equation}
where $x$ is the estimated PET and $y$ is the ground truth PET. $\mu_i$ is the mean of image $i$, $\sigma_i$ is the variance of image $i$ and $\sigma _{xy}$ is the co-variance of images $x$ and $y$. $C_{1}$ and $C_{2}$ are empirically found constants in order to best perceive the structure of the estimated PET with respect to the ground truth.

Table \ref{tab:metrics} shows the performance of both the approaches using all three quantitative metrics. The high PSNR~$(68.88)$ and SSIM~$(0.82)$ and low MAE~(0.0422) indicate that our proposed global method outperforms the local approach in all metrics, demonstrating the superiority of the architecture.

\begin{table}[!htbp]
  \begin{center}
    \scalebox{1.0}{
    \def\arraystretch{2.0}
    \begin{tabular}{|c|c|c|c|} 
    \hline
       Method & \textbf{SSIM} & \textbf{MAE} & \textbf{PSNR}\\
      \hline
      Li et. al\cite{patch} & $0.5419 \pm 0.044$ & $0.0862 \pm 0.0003$ & $58.29 \pm 1.337$  \\ \hline 
      Proposed Method & \textbf{0.8211 $\pm$ 0.015} & \textbf{0.0422 $\pm$ 0.006} & \textbf{68.88 $\pm$ 1.010} \\ \hline 
    \end{tabular}}
    \bigskip
    \caption{Comparison of \cite{patch} with our proposed global approach for estimation of PET from MRI. For SSIM and PSNR metrics, the higher the value, better the estimation and for MAE metric, the lower the value, better the estimation. }
    \label{tab:metrics}
  \end{center}
  \vspace{-10mm}
\end{table}

\begin{figure*}[!h]
  \centering
  \includegraphics[width=\textwidth]{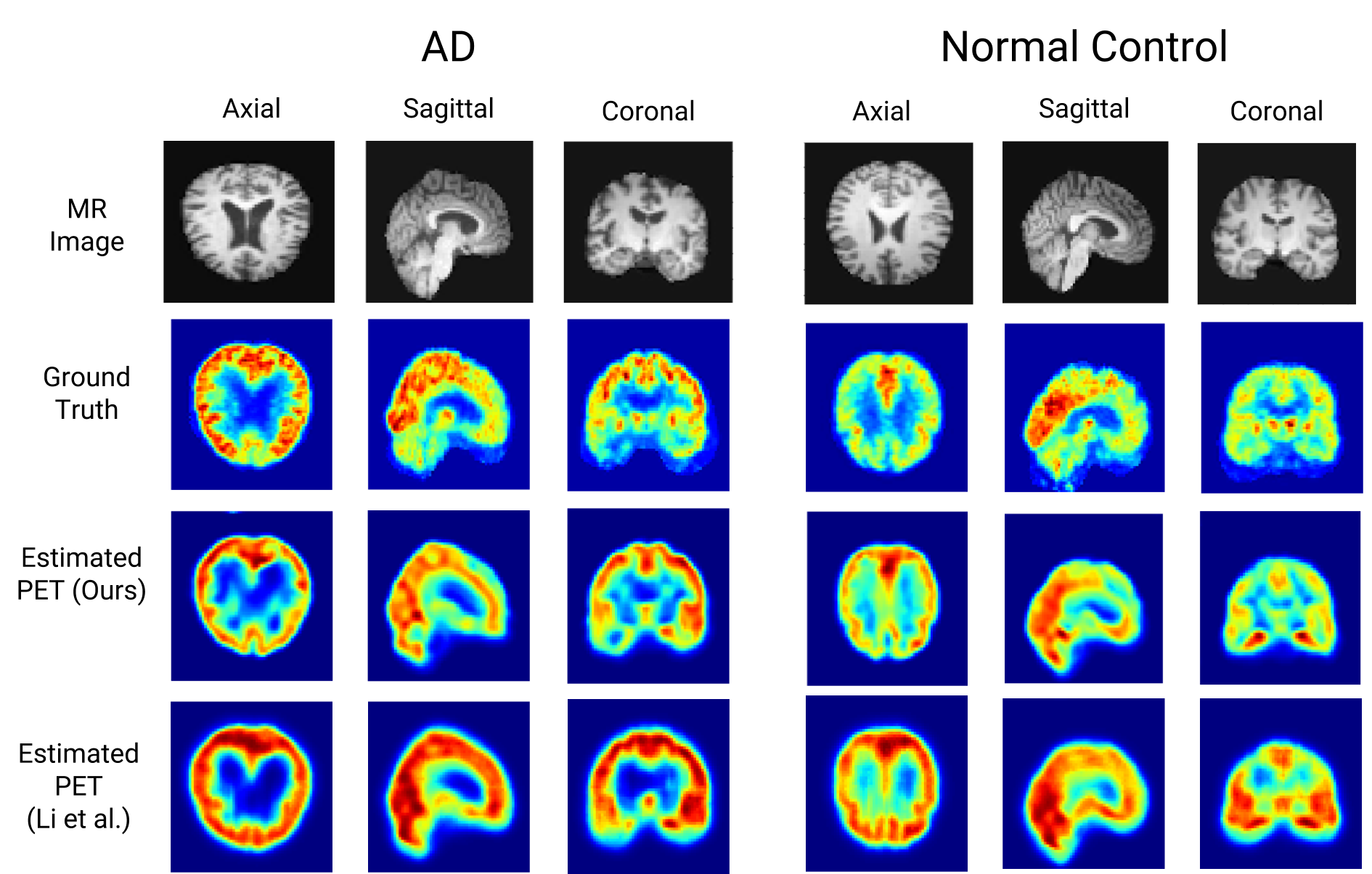}
  \caption{Qualitative comparison of predicted PET scans with their corresponding true scans for 2 subjects -- one from each group: Normal and AD in axial (right) and saggital views (left).} 
   \label{fig:example}
\vspace{-7mm}
\end{figure*}

For a more qualitative analysis of the results, samples of PET scans estimated by both the approaches along with original PET scans are displayed in figure \ref{fig:example}. The results clearly illustrate higher level of similarity between the PET scans estimated by the proposed method and their respective ground truth scans as compared to the alternative. The images further corroborate  numerical results presented above and demonstrate the potential of the proposed approach to learn features corresponding to Normals and AD.

\subsection{Impact of the Proposed Method on Alzheimer's Classification}
To evaluate the effectiveness of the estimated PET scans, we performed classification of Alzheimer's Disease on the ADNI \cite{adni} dataset using reconstructed data. As 9 fold cross-validation was used as part of the cross-modality estimation procedure, the same setting was used to perform the classification task. To ensure consistency with \cite{patch}, we used $\ell2$-norm regularized logistic regression classifiers for both methods. The results of classification task using both the patch based method \cite{patch} and the proposed U-Net based method are shown in Table \ref{tab:classification}. As expected, we observe that the joint classification accuracy of MRI+Synthesized PET results in higher accuracy than stand alone MRI based classification due to the complementary nature of the features extracted and utilized from both modalities. The joint accuracy for \cite{patch} is less than that of only MRI might be due to considering only local correlations to generate the PET which is actually misclassifying the image. A paired sample \textit{t}-test revealed marginally significant improvement ($p = 0.06$) in classification accuracy using MRI and synthesized PET images.
\begin{table}[!htbp]
  \begin{center}
    
    \scalebox{1.0}{
    \def\arraystretch{2.0}
    \begin{tabular}{|c|c|c|c|c|} 
    \hline
       Method & \textbf{MRI} & \textbf{PET} & \textbf{PET-Synthesis} & \textbf{MRI+PET-Synthesis}\\
      \hline
      Li et. al\cite{patch} & 70.18 $\pm$ 8.37 & 80.80 $\pm$ 7.95 & 60.33 $\pm$ 6.14 & 65.34 $\pm$ 5.43 \\ \hline 
      Proposed Method & 70.18 $\pm$ 8.37 & 80.80 $\pm$ 7.95 & \textbf{69.95 $\pm$ 5.59} & \textbf{74.43 $\pm$ 3.32} \\ \hline 
    \end{tabular}}
    \bigskip
    \caption{Accuracies on the binary classification task of AD vs Normal by the proposed method and the patch based method \cite{patch}. The second and the third columns (MRI \& PET) have the same set of values as they are the classification accuracies on the original data and do not depend upon the method.}
    \label{tab:classification}
  \end{center}
  \vspace{-12mm}
\end{table}

\section{Discussion}
We hypothesized that the correlation present between MRI and PET images is not local \cite{multimodal}. Ideally, a fully connected network~(FCN) would be more appropriate to model global correlations between MRI and PET scans. However, due to large number of parameters and limited number of data samples, FCN produced relatively smoother estimates of PET scans. We initially experimented by training a supervised autoencoder architecture having two layers with each hidden layer having 500 and 300 hidden units. The number of parameters in this architecture in first layer itself rise to $131$ million which increases the chances of overfitting given such less number of samples in medical images. With simple convolutional neural network the number of parameters reduces but the context it captures remains local. But U-Net is a convolutional encoder-decoder architecture with skip connections which helps us capture global correlations while keeping the number of parameters low.

We used binary crossentropy as loss function to train models instead of mean squared error~(MSE) as it was generating jittery outputs, whereas the outputs were smooth in the former case. The choice of classifier was made to be consistent with \cite{patch}, which could be improved by using deep learning based classifiers. 

The estimated PET scans were evaluated against three global metrics as discussed above. Additionally, we tried to evaluate the quality of PET scans for few regions-of-interests~(ROI) locally responsible for AD. For this, we parcellated estimated PET into 120 regions using AAL \cite{parce}. From these regions, we select few important regions highlighting AD as described in \cite{jack1}. Table \ref{tab:discussion1} highlights mean MAE and PSNR values for these regions. The range of MAE and PSNR values across all regions is similar, indicating the quality of generated scans locally.  

\begin{table}[!h]
  \begin{center}
    \scalebox{0.95}{
    \def\arraystretch{1.2}
    \begin{tabular}{|l|c|c|} 
    \hline
       \textbf{Name} & \textbf{MAE} & \textbf{PSNR} \\
      \hline
      Hippocampus  & 0.22  &  60.30 \\ \hline 
      Para Hippocampus & 0.22 & 60.39  \\ \hline
      Posterior Cingulate & 0.21 & 60.58  \\ \hline
      Precuneus & 0.18 & 61.52  \\ \hline
      Anterior Cingulate & 0.16 & 61.14  \\ \hline
      Orbito Frontal & 0.18 & 61.33  \\ \hline
    \end{tabular}}
    \bigskip
    \caption{Quantitative Evaluation of metrics for few ROIs significant for Alzhiemer's   }
    \label{tab:discussion1}
  \end{center}
\end{table}
\vspace{-15mm}

\section{Conclusion}

We have explored U-Net architecture to estimate PET modality which when used alongwith MRI improves classification accuracy over the state-of-the-art. A 3D architecture which takes full MRI volume as an input and generates a corresponding PET scan in one pass is used to perform the cross-modal estimation. 
The presence of skip connections allows the model to capture both non-linear and non-local correlations in an encoder-decoder setting. We have demonstrated the applicability of generated scans via performing multi-modality classification using both original MR and synthetic PET scan. The increased joint classification accuracies imply that synthetic data can be used in cases where capturing PET scans is not feasible. It can also be used as a missing data method for estimating PET scans that have been omitted for various reasons. We plan to extend this by making estimations more strong using adversarial training with the same U-Net architecture.

%
%
\bibliographystyle{splncs04}
\bibliography{samplepaper}

\end{document}